# A simple method for decision making in RoboCup soccer simulation 3D environment

# Un método simple para la toma de decisiones en ambientes 3D de simulación de fútbol RoboCup

Khashayar Niki Maleki[1], Mohammad Hadi Valipour[1], Roohollah Yeylaghi Ashrafi[1], Sadegh Mokari[1], M. R. Jamali[1,2], Caro Lucas[2]

1. Department of Electrical and Computer Engineering, Shahid Rajaee University, Tehran, Iran
2. Control and Intelligent Processing Center of Excellence, School of Electrical and Computer Engineering, University of Tehran, Tehran, Iran

{kh.niki, m.h.valipour, r.yeylaghi, s.mokari}@sru.ac.ir, m.jamali@ece.ut.ac.ir, lucas@ut.ac.ir

**Abstract**—In this paper new hierarchical hybrid fuzzy-crisp methods for decision making and action selection of an agent in soccer simulation 3D environment are presented. First, the skills of an agent are introduced, implemented and classified in two layers, the basic-skills and the high-level skills. In the second layer, a two-phase mechanism for decision making is introduced. In phase one, some useful methods are implemented which check the agent's situation for performing required skills. In the next phase, the team strategy, team formation, agent's role and the agent's positioning system are introduced. A fuzzy logical approach is employed to recognize the team strategy and furthermore to tell the player the best position to move. At last, we comprised our implemented algorithm in the Robocup Soccer Simulation 3D environment and results showed the efficiency of the introduced methodology.

**Resumen**—En este artículo se presenta un nuevo método difuso híbrido para la toma de decisiones y selección de acciones de un agente en un ambiente de simulación de fútbol 3D. Primero, se introducen, implementan y clasifican las competencias del agente en dos capas: competencias básicas y competencias de nivel superior. En la segunda capa se introduce un mecanismo de dos fases para la toma de decisiones. En la primera fase, se aplican algunos métodos útiles que permiten verificar la situación del agente para la ejecución de las competencias requeridas. En la siguiente fase, se adicionan la estrategia y formación del equipo, y el rol y sistema de posición de los agentes. Se emplea la lógica difusa para reconocer la estrategia de equipo y además indicarle al jugador, la mejor posición para moverse. Por último, se incluye el algoritmo implementado en el ambiente 3D de Simulación de Fútbol RoboCup y los resultados que demuestran la eficiencia de la metodología introducida.

**Palabras Clave**—Sistemas multi-agente, Aprendizaje de máquina, Inteligencia artificial, Refuerzo del aprendizaje, Lógica difusa, Refuerzo del aprendizaje difuso, Simulación de fútbol RoboCup.

## I. INTRODUCTION

Robocup is continuing AI research initiative that uses the game of soccer as a unifying and motivating domain [1, 2]. The Robocup simulation competition pits teams of 11 independently-controlled autonomous agents against each other in Robocup simulator, or Soccer Server, a real-time, dynamic environment [3]. The only Robocup soccer simulator used to be «2D» for years, many researches in AI have performed in that full challenging environment for functionality of the developed algorithm in a multi-agent environment [4]. But as it has been set by RoboCup Federation [RoboCup, 2004], the ultimate goal can be expressed by several discussion. «By mid-21st century, a team of fully autonomous humanoid robot soccer players shall win a soccer game, complying with the official rules of the FIFA, against the winner of the most recent world cup for human players.» [Kitano & Asada, 1998] So there was a need for a more realistic platform rather than the previous 2D environment which is more familiar with real soccer game. That leaded to Robocup 3D Soccer Simulator.

As the optimal scoring problem is well suited for Machine Learning (ML) techniques [5], Nowadays many powerful methods with different roots, has been introduced in ML such as: neural networks [5], genetic algorithms [6], genetic programming [6], fuzzy logic [7], coordination graphs [8,9] and also many hybrid approaches as a combination of some aforementioned ones [4] ML techniques have been applied to a variety of problems including data mining, pattern recognition, classification problems, e.g. road condition classifier [Ferdowsizadeh, 2004], adaptive control, robot control, combinatorial optimization and game playing. There are lots of publications in applying these methods in RoboCup Soccer Simulated teams mostly in 2D (and NOT 3D) environment [1, 2, 3, 4, 5, 6, 8, and 9] (there are much more than referenced researches). It may have different reasons, but having problem with low-level skills that leads to disability in controlling agents, complicated dynamics which makes predictions not to work well and a vast range of performing soccer skills in comparison with 2D simulated environment for sure are the most important effecting factors. There are new features and also limitations in this environment which make some distinctions in decision making process. (For more detailed information you can refer to [10, 11])

In this paper a new methodology for decision making in RoboCup soccer simulated 3D environment with all abovementioned problems consideration is introduced. Implementation of real soccer skills in two layers and utilizing both fuzzy and non-fuzzy algorithms in different layers of decision making are the essential keys for the accomplishment of this system. Section two reviews the state of art of this methodology in order to apply it in decision making process. In this section the basic skills and their functionality and also some of soccer (high-level) skills are introduced and classified. The first layer of decision making process is discussed in section three and the second layer in section four. Section five comprises our results which are implemented on Scorpius Soccer Simulation Team[1] and finally section six concludes the paper.

## II. LAYERS OF SKILLS AND DECISION MAKING PROCESS

In the proposed methodology some applicable skills are introduced and the decision making policy is developed by considering the features and limitations of this environment. The skills are classified in two layers, in the first layer there are the simple actions which are already implemented by server

(basic skills) and in the second layer the actions are more complicated and sometimes a combination of «basic skills» are used (high-level skills). Besides, the decision making process has two steps; first step considers the agent's abilities of performing a high-level skill according to his circumstances and the second step considers the agent positioning and choosing the best action regarding to the first step results.

### A. Basic skills

The basic skills are defined as the actions, which are already implemented by server [10, 11]. They can be employed by sending the proper commands to soccer server[2]. These commands are:

· *A(drive x y z)*: moves the agent by applying the force vector (x, y, z) to center of it.
· *A(kick alpha f)*: agent kicks the ball by applying the force *f* with the angle *alpha* to it if the ball is in kickable distance.
· *A(pantilt angle1 angle2)*: This command changes the view direction of an agent where «angle1»and «angle2» are changes (in degrees) of the pan and tilt angle, respectively.
· *A(say(«say message here»))*:sends a message to all the players that are located in 50 meter from sender.
· *A(catch)*: (for goal keeper only) holds and freezes the ball if the ball is in the catch able area.

(See [10, 12] for more detailed information about the commands)

### B. High level skills

High-level skills are those in which the world model[3] information and the basic skills are being applied. These skills consist of the actions with ball like: pass, shoot, dribble, etc. and actions without ball like: mark, find object and information broadcasting.

#### 1) Actions with ball

To perform these actions, basically we need some information about the ball treats, when a force with a particular angle is applied, considering the environment parameters (e.g. air force, friction, collision, etc). A number of complicated mechanical formulas could help to predict the ball movements (Figure 1 shows the ball predicted motion in comparison with the real ball position information given from the monitor[4]). Some of the most applicable skills with ball are as follows:

---

2. These skills are implemented in rcsoccersim3D_0.5.5
3. World Model is a data bank in which the information about the environment is stored
4. We wrote a program to parse the monitor log file. We assume that this information is the most accurate one we may have from what really happens in server.

· *Shoot*

Shooting for sure is the most important skill in soccer and all other skills like pass, dribble, etc are based on it. A player in this environment can only kick the ball in front of himself, so he needs to get behind ball in a correct position to shoot the ball in his desired direction (see Figure 4).

· *Pass*

The agent kicks the ball so that the other teammate can receive it. This can be a simple definition for the skill «pass». As this skill can be used in different situations, the *type* is defined for it. We have three kinds of pass:

1. *Secure pass*. The player is completely sure that this pass will arrive to the player he wishes. The most popular usage of this type of pass is when the ball is in danger zone[5].
2. *Normal pass*: the probability of success in pass is more than its failure. Mostly used in the middle of the field.
3. *Risky pass*: In this case, the probability of ball arrival exists, but the possibility of failure is more than its success. This type is used in order to create a good situation for team.

· *Dribble (run with ball)*

The agent usually uses the skill «dribble» when he owns the ball in an almost free space and can't find the other agents with better positions or can't pass them the ball (see Figure 7b).

· *Clear ball*

This is an action that agent chooses to do, when he owns the ball and can perform no other action or the player is in dangerous situation, mostly happens in *defense*. Depending on the occurred circumstance, the agent may kick the ball out of the field, toward the opponent goal or other positions.

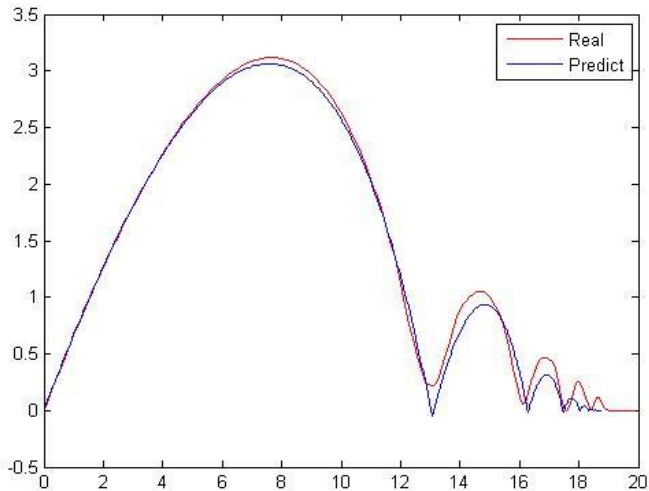

Figure 1. The comparison between the mechanical formula used for ball motion prediction and the real data given from the monitor.

5. Danger zone is defined in fuzzy member function (see Fig. 4)

2) Actions without ball

These skills make agents to be arranged in positions so that they would have the most chance to create opportunities for team or to get the opponents opportunities.

· *Mark*

Mark skill approaches two purposes:

1. Not to let the ball reaches the opponents (mark player).
2. Not to let the opponents shoot to their desired position (mark ball).

According to the purpose the player gets near to the opponent up to the *MarkSecureDistance* and marks him.

· *Pan-tilt (object-finding skill)*

The agent uses this skill to find an object and/or to update his world model. There are two different conditions:

1. The agent sees an object but wants to keep it in the center of his vision not to let that object gets out sight easily. This usually happens for ball because its location varies very fast and easily may get out of the agent's sight (Figure 2).
2. The agent doesn't see the object; in this case he pans with *MaxPanAngle* with the direction in which the last time that object was seen.

· *Say (information broad casting or alerting skill)*

This skill is being used for alerting the agents and also to update their world models. Depending on how many characters per cycle an agent can talk, he can use them in order to update the other agents' world models. One of the most recommended usages of this skill is utilized in mark. When the defenders «*mark ball*» the opponents they do not see the opponents which are marked. So they need to change their view in each few cycles, but when the ball gets near they may focus on ball and forget about the agent. In this case or similar events we can use say skill to update and also alert the agents. Figure 2 shows this event.

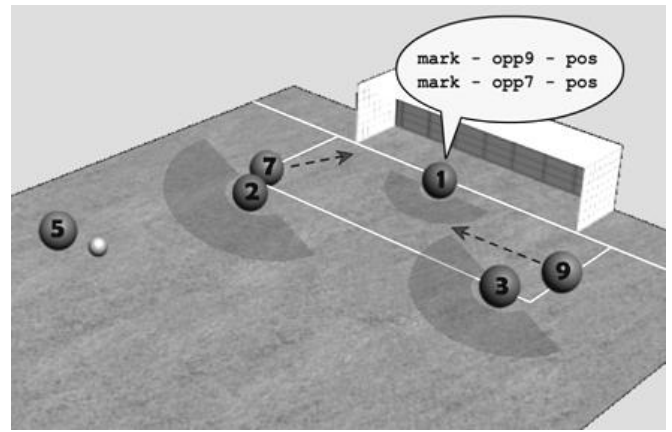

Figure 2. The goal keeper says a message to update the defenders world model. They can't see the opponents' displacement behind them.

### III. DECISION MAKING PHASE 1

In this phase, the agent ability of performing an action according to the environment situation (e.g. opponents' positions/speed, ball position/speed and their predicted states) is tested by Decision Makers (DM) to confirm if that action can be done by the agent in that condition. Some of these decision makers are explained below:

· *DM for shooting to a position*

To determine if an agent can shoot to a position or not, first the agent calculates the minimum degree, for shooting, if the degree could be found less than *MaxKickDegree* then tries to find an angle between min degree and *MaxKickDegree* and a force less than *MaxKickForce*. If the angle and a force found with these properties the DM returns *true*, otherwise it returns *false*.

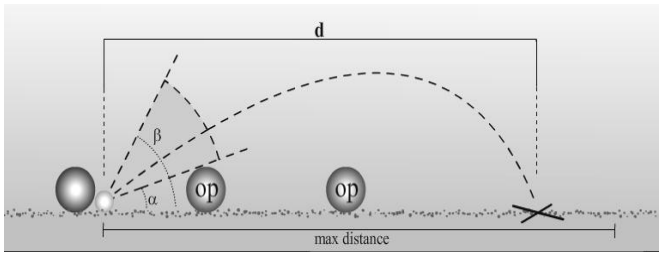

Figure 3. Shooting to a position requires minimum angle á to be less than (*Max Kick Angle*) and minimum distance *d* less than *(Max Kick Distance)*

· *DM for Shoot to goal*

First, the agent quantizes the goal, to *n* discrete positions. For each position first checks the *Shoot to position* conditions, if the result is true then checks the following condition:

Let Tb be the time takes ball to meet the target with the maximum speed, and Tr be the rotation time for the ball controller to adjust it's position beside the ball. Tg represents the time takes goalie to catch the ball (Figure4). Having calculated the above three parameters we define $\Delta t$ as following [7]:

$$\Delta t = Tg - (Tb + Tr) \qquad (1)$$

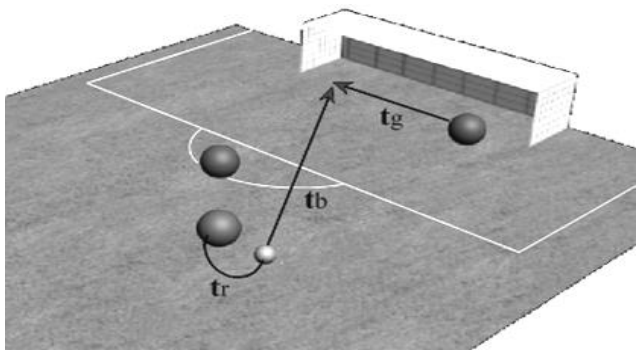

Figure 4. The time intervals needed for prediction calculation applied in shoot-to-goal decision maker.

The sign of $\Delta t$ shows if the agent can shoot to that position or not.

### *DM for pass*

This DM gets the ball position and the pass receiver as input parameters. to know where to use which type we implemented a fuzzy algorithm. This fuzzy algorithm according to the ball position and the player situation tells us which type is more proper and also gives the *Max Allowed Pass Error*.

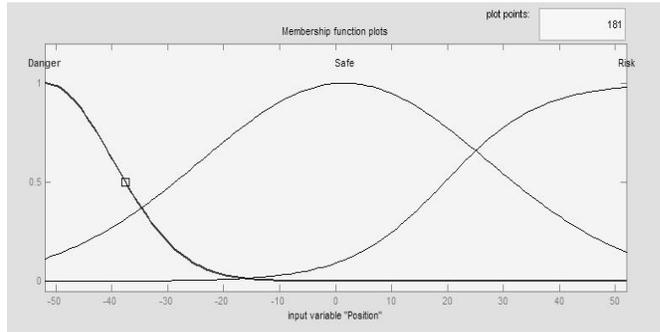

a

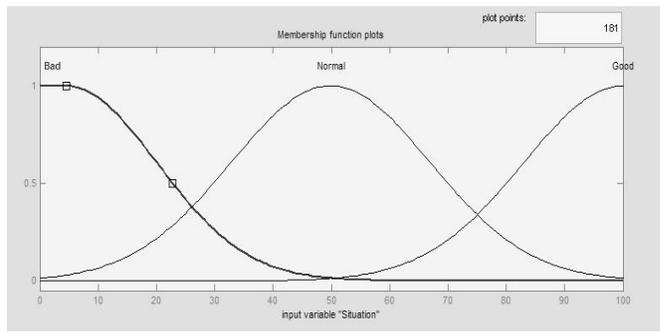

b

Figure 5. Fuzzy input member functions; a. ball position input variable b. pass receiver situation input variable

Fuzzy rule base for pass is as follows:

1. If (position is Danger) then (pass type is Secure).
2. If (position is Safe) and (situation is Bad) then (pass type is Secure).
3. If (position is Safe) and (situation is Normal) then (pass type is Normal).
4. If (position is Safe) and (situation is Good) then (pass type is Risky).
5. If (position is Risk) and (situation is Bad) then (pass type is Secure).
6. If (position is Risk) and (situation is Normal) then (pass type is Normal).
7. If (position is Risk) and (situation is Good) then (pass type is Risky).

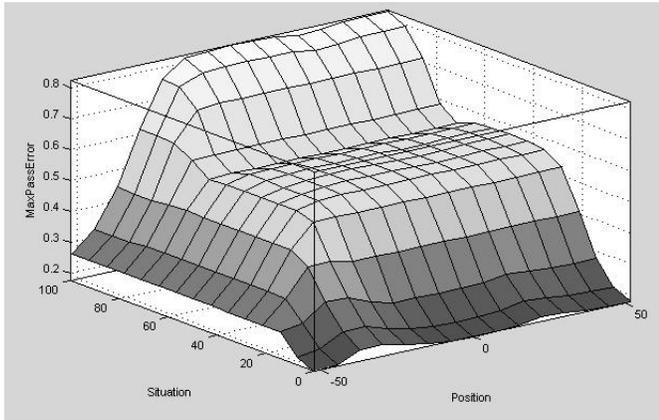

Figure 6. The relationship between ball position-pass receiver situation and *Max Pass Error* according to the fuzzy algorithm.

After the pass type and *Max Pass Error* became determined, agent checks his ability for performing this type of pass.

First the agent checks if he *can shoot to* the pass position or not then the amount of risk will be calculated. The following factors have to be checked in order to calculate the risk of pass:

1. Opponents around the ball
2. The relative agent position with ball.
3. The max error for getting position behind the ball
4. The position of opponents around the player we want to pass to, and their distances with him.
5. The ball path

Let *t1* be the time takes player to get position behind ball. It depends on factors 1,2, and 3 which vary with pass type, and *t2* be the time takes for opponents to reach the ball, the player *can pass* when *t1* is less than *t2*.(See Figure 7.a)

Let $t(i)$ be the time takes for opponent $i$ to intercept the ball considering the ball height from the field at position $p(i)$ with error $E1$, and $t1(i)$ is the time takes for ball to reach $p(i)$ with error $E2$, where $E1$ and $E2$ vary with pass type. If $t1(i)<t(i)$ for all $i$ opponents then player *can pass*. And the last factor is when the target player, gets the ball. He must have *Secure Time* (t5 in Figure 7.a) to control the ball, which varies with pass type. Now with the amount of risk, the player can determine whether he can pass with requested type or not.

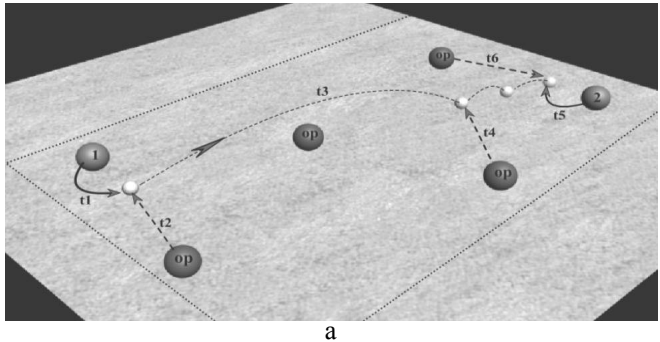

a

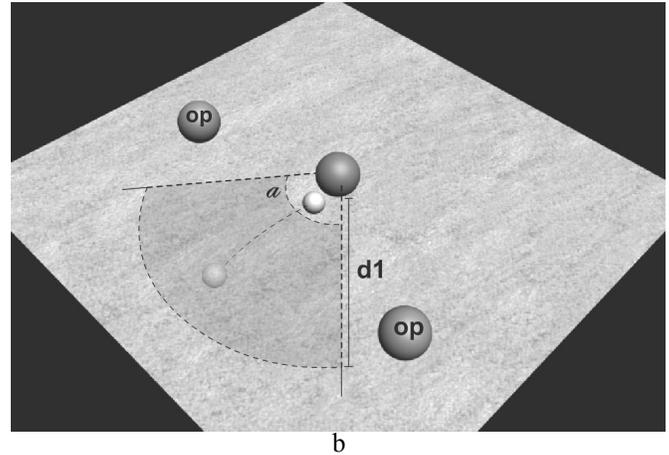

b

Figure 7. a. Player 1 wants to pass to player 2,  b. d1 is the Secure Dribble Distance and á is the Secure dribble angle

· *DM for Dribble*
To determine if it is safe enough for player to dribble (run with ball)

First, the player draws a cone with *SecureDribbleAngle* in front of himself, Then he checks If there is no opponent with distance less than or equal with *SecureDribbleDistance* inside the cone subsequently he predicts the ball position after kick when the ball speed is less than *MaxBallControllableSpeed* let it take t1, next he predicts the opponent players plus the agent himself positions after t1, assuming that they move to the ball predicted position with their *MaxSpeed*. If the nearest player to the ball after t1 is the agent himself then the DM returns true, otherwise returns false (Figure 7.b).

· *Find Opponent to mark (DM for marking)*
The player searches his strategic area (see section V for more details) for *not marked* opponents, if there were more than one opponent, and then checks the possibility of marking each opponent. This possibility may have the same factors we represented in *can pass* function. At last, player can determine the best player to mark.

### IV. DECISION MAKING PHASE 2

The major issues we have addressed in this phase are the static assignment of roles and dynamic team strategy. We adopted a formation/role system similar to one described in [13, 14, and 15] each formation contains:

- *Formation name* : Like real soccer team formations (e.g. 4_4_2)
- *Strategic area*: The area in which the player is mostly supposed to be.
- *Center of strategic area*: also known as the home position
- *Player role*: we introduced 4 applicable roles for agents: goal keeper, defender, half backer and attacker.

In this methodology the player role is statically assigned to the player according to his player number, but the team strategy varies with different factors. The most important factor is the ball position. According to this changes in strategy the strategic area of the player, changes. To select the appropriate strategy we developed a fuzzy algorithm (Figure 8). In addition this algorithm helps an agent to find out the proper distance with ball according to his strategic area (Figure 9.).

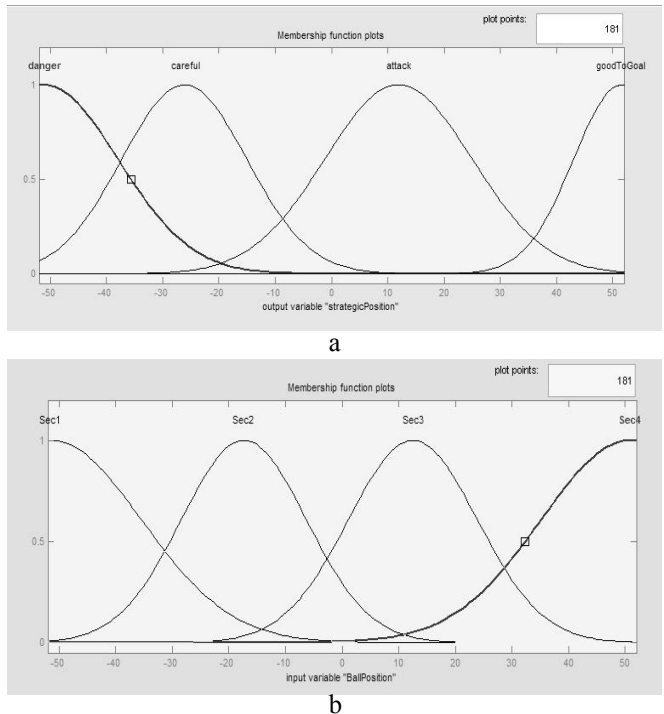

a

b

Figure 8. Fuzzy member function for selecting strategy; a. output variable *strategy* b. input variable *ball Position*

Fuzzy rule base for strategy is as follows:

1. If (BallPosition is Sec1) then (strategy is Danger)
2. If (BallPosition is Sec2) then (strategy is Careful)
3. If (BallPosition is Sec3) then (strategy is Attack)
4. If (BallPosition is Sec4) then (strategy is GoodToGoal)

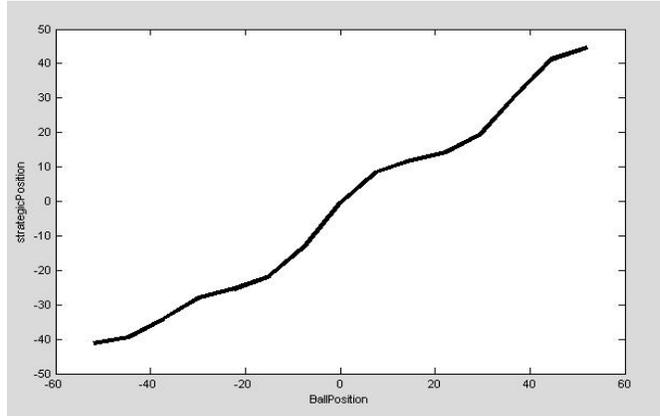

Figure 9. The relationship between ball position and a *half back* player strategic position in fuzzy algorithm.

Depending on the agent's team or the opponents' team owns the ball (see sec. IV.*B*.) the output variable strategy (Figure 8.a) of fuzzy function may change. The last remaining condition is when the agent owns the ball (see sec. IV.*A*), in this state agent uses the phase 1 decision making results to perform an appropriate operation (Figure 10).

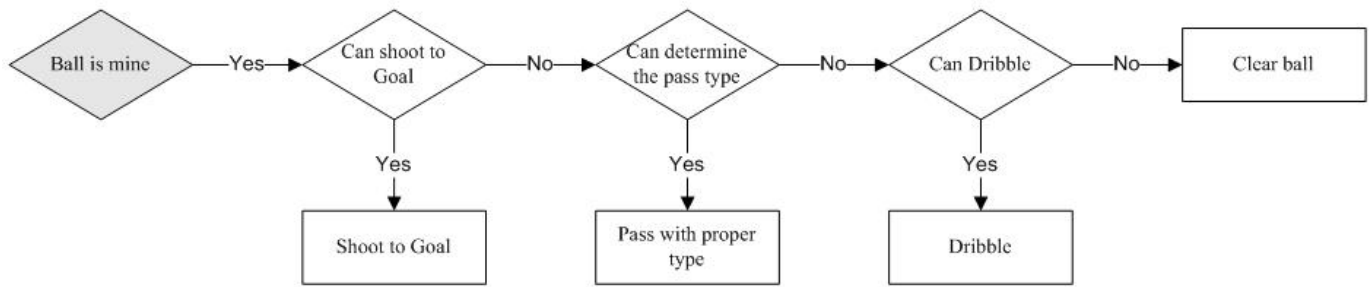

Figure 10. The diagram for action selection when players own the ball

### A. Who is the nearest player to ball?

A function is implemented that predicts the ball position after its speed is less than or equal with *MaxBallControllableSpeed*, Then returns the player who is the closest to that position.

### B. Which team owns the ball?

An other function is implemented to check if the ball speed is less than *MaxBallControllableSpeed* then calls the *Nearest Player to ball* function, if the player team side which this function returns, was the same of ours, then this function returns 1 other wise returns 0, But if the ball speed is more than the value mentioned above, then this function returns -1, that means we can not determine whether ball is ours or not.

## V. RESULTS

We implemented this methodology on Scorpius Soccer Simulation Team. As there are not many source codes and/or binaries of 3D teams adoptive with latest changes of server[6], we decided to compare this team with its previous version[7]. The results showed the success of this methodology; the team performance in coordination and collaboration highly improved, in fact the players switched their strategic area smoothly as the team strategy changed in a reasonable manner, the agents carried out the high-level skills much more efficiently and at last the final results enhanced significantly (Table 1. Shows the final match results).

## VI. CONCLUSION

In this paper, a new method for decision making in RoboCup soccer 3D simulation environment has been proposed. First an identification applied to the Robocup Soccer 3D environment which led to mechanical formulas, then soccer skills were introduced and classified in different layers then a two-phase mechanism for decision making is presented, in this mechanism both fuzzy and non-fuzzy algorithms are applied and finally the proposed methodology implemented on a Robocup soccer simulation team and the results showed the efficiency of this methodology.

## ACKNOWLEDGEMENT

Authors express thanks to Mr. S. Mahdi Hosseini from the ECE department, University of Theran, for the ideas, which ones were used at writing article.

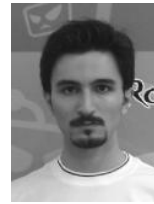

**Khashayar Niki Maleki** is B.Sc. student in Computer Hardware Engineering in Department of Electrical and Computer Engineering, Shahid Rajaee University, Tehran, Iran. He is a member of Scorpius Simulation Team, which has participated in many Robocup competitions. He is IEEE/ACM student member. His interested research areas are multi agent systems, reinforcement learning and System Identification and programming for embedded systems.

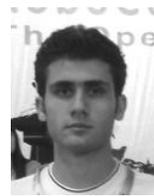

**Mohammad Hadi Valipour** received the B.Sc. degree in Information Technology Engineering from Department of Electrical and Computer Engineering, Shahid Rajaee University, Tehran, Iran in 2008. He is a member of Scorpius Simulation Team, which has participated in many Robocup competitions. Also he is IEEE/ACM student member. His interested research areas are software architecture, agent oriented programming, communication systems, and multi agent systems.

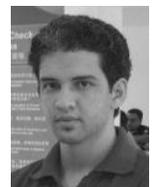

**Sadegh Mokari** is B.Sc. student in Computer Hardware Engineering in Department of Electrical and Computer Engineering, Shahid Rajaee University, Tehran, Iran. He is a member of Scorpius Simulation Team, which have participated in many Robocup competitions. He is ACM student member. His interested research areas are multi agent systems and artificial neural networks.

---

6. The simulation environment used for experimental results is :rcsoccersim3D.0.5.5

7. This version was so similar to which placed 6th in Iran Open 2006 International Robocup Competitions.

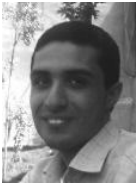
**Roohollah Yeylaghi Ashrafi** is B.Sc. student in Computer Hardware Engineering in Department of Electrical and Computer Engineering, Shahid Rajaee University, Tehran, Iran. He is a member of Scorpius Simulation Team, which have participated in many Robocup competitions. He is interested in open source projects and multi agent systems.

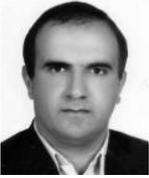
**Mohammad Reza Jamali** received the B.Sc. degree in Computer Hardware Engineering from Amirkabir University of Technology in 2001. He got the M.Sc. degree from Shiraz University in the field of Artificial Intelligence in 2003. He is a Ph.D. candidate since 2004 under the supervision of Professor Lucas in Centre of Excellence for Control and Intelligent Processing at the Department of Electrical and Computer Engineering, University of Tehran, Iran. He is interested in evolutionary algorithms, multi agent systems and design methodologies in control and automation systems. Also he is the leader of Scorpius Simulation team since 2005.

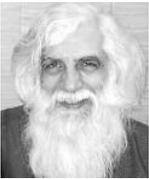
**Caro Lucas** received the M.Sc. degree from the University of Tehran, Iran, in 1973, and the Ph.D. degree from the University of California, Berkeley, in 1976. He is a Professor of Centre of Excellence for Control and Intelligent Processing at the Department of Electrical and Computer Engineering, University of Tehran, Iran, as well as a Researcher at the School of Intelligent Systems (SIS), Institute for Studies in Theoretical Physics and Mathematics (IPM), Tehran, Iran. He has served as the Director of SIS (1993-1997), Chairman of the ECE Department at the University of Tehran (1986-1988), Managing Editor of the Memories of the Engineering Faculty, University of Tehran (1979-1991), Reviewer of Mathematical Reviewers (since 1987), Associate Editor of Journal of Intelligent and Fuzzy systems (1992-1999), and Chairman of the IEEE, Iran Section (1990-1992). He was also a Visiting Associate Professor at the University of Toronto (summer, 1989-1990), University of California, Berkeley (1988-1989), an Assistant Professor at Garyounis University (1984-1985), University of California at Los Angeles (1975-1976), a Senior Researcher at the International Centre for Theoretical Physics and the International Centre for Genetic Engineering and Biotechnology, both in Trieste, Italy, the Institute of Applied Mathematics Chinese Academy of Sciences, Harbin Institute of Electrical Technology, a Research Associate at the Manufacturing Research Corporation of Ontario, and a Research Assistant at the Electronic Research Laboratory, University of California, Berkeley. He is the holder of Patent on «Speaker Independent Farsi Isolated Word Neurorecognizer». His research interests include biological computing, computational intelligence, uncertain systems, intelligent control, neural networks, multi-agent systems, data mining, business intelligence, financial modeling and knowledge management. Professor Lucas has served as the Chairman of several International Conferences. He was the Founder of the SIS and has assisted in founding several new research organizations and engineering disciplines in Iran. He is the recipient of several research grants at the University of Tehran and SIS.

## FE DE ERRATAS

En la edición anterior de la Revista Avances en Sistemas e Informática, Volumen 5 número 3 de Diciembre de 2008, no se presentó la Tabla 1: Final results, en la página 115, perteneciente al artículo:

# Un método simple para la toma de decisiones en ambientes 3D de simulación de fútbol RoboCup

Khashayar Niki Maleki
Mohammad Hadi Valipour
Roohollah Yeylaghi Ashrafi
Sadegh Mokari
M. R. Jamali
Caro Lucas

La Tabla 1 completa se presenta a continuación:

**Table 1.** Final results

| Match | Goals scored by team A [a] | Goals scored by team B |
|---|---|---|
| 1 | 2 | 0 |
| 2 | 3 | 0 |
| 3 | 3 | 1 |
| 4 | 3 | 1 |
| 5 | 2 | 1 |
| Average | 2.6 | 0.6 |

a: We applied the methodology to the team A